\documentclass{article}
\usepackage[preprint]{neurips_2024}

\usepackage[utf8]{inputenc} %
\usepackage[T1]{fontenc}    %
\usepackage{hyperref}
\usepackage{url}            %
\usepackage{booktabs}       %
\usepackage{natbib}

\usepackage{multicol}
\usepackage{xcolor}   
\usepackage{multirow}
\usepackage{microtype}      %
\usepackage{enumitem}
\usepackage{tabularx}
\usepackage{graphicx}       %
\usepackage{caption} 
\usepackage{float}
\usepackage{adjustbox}
\usepackage{amsmath, amsfonts, amssymb, bbm}

\newcommand{\tabincell}[2]{\begin{tabular}{@{}#1@{}}#2\end{tabular}}

\title{Yi-Lightning Technical Report}

\author{
  \textbf{01.AI} %
}
\begin{document}
\maketitle

\begin{abstract}

This technical report presents \textbf{Yi-Lightning}, our latest flagship large language model (LLM). 
It achieves exceptional performance, ranking \textbf{6th} overall on Chatbot Arena, with particularly strong results (2nd to 4th place) in specialized categories including Chinese, Math, Coding, and Hard Prompts.
Yi-Lightning leverages an enhanced Mixture-of-Experts (MoE) architecture, featuring advanced expert segmentation and routing mechanisms coupled with optimized KV-caching techniques.
Our development process encompasses comprehensive pre-training, supervised fine-tuning (SFT), and reinforcement learning from human feedback (RLHF), where we devise deliberate strategies for multi-stage training, synthetic data construction, and reward modeling.
Furthermore, we implement RAISE (Responsible AI Safety Engine), a four-component framework to address safety issues across pre-training, post-training, and serving phases.
Empowered by our scalable super-computing infrastructure, all these innovations substantially reduce training, deployment and inference costs while maintaining high-performance standards.
With further evaluations on public academic benchmarks, Yi-Lightning demonstrates competitive performance against top-tier LLMs, while we observe a notable disparity between traditional, static benchmark results and real-world, dynamic human preferences.
This observation prompts a critical reassessment of conventional benchmarks' utility in guiding the development of more intelligent and powerful AI systems for practical applications.
Yi-Lightning is now available through our developer platform at \url{https://platform.lingyiwanwu.com}.

\end{abstract}

\begin{figure}[h]
  \centering
  \vspace{1mm}
  \includegraphics[width=0.55\linewidth]{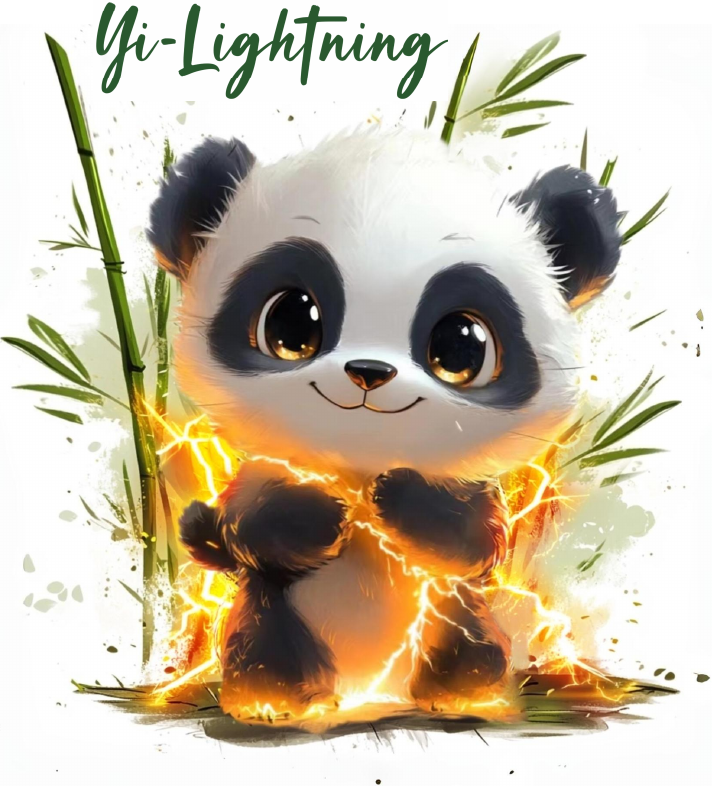}
  \caption*{\textcolor{white}{Yi-Lightning logo.}}
  \vspace{-3mm}
  \label{fig:logo}
\end{figure}

\clearpage

\tableofcontents

\clearpage

\section{Introduction}

\begin{figure}[!h]
    \centering
    \vspace{1mm}
    \includegraphics[width=0.85\linewidth]{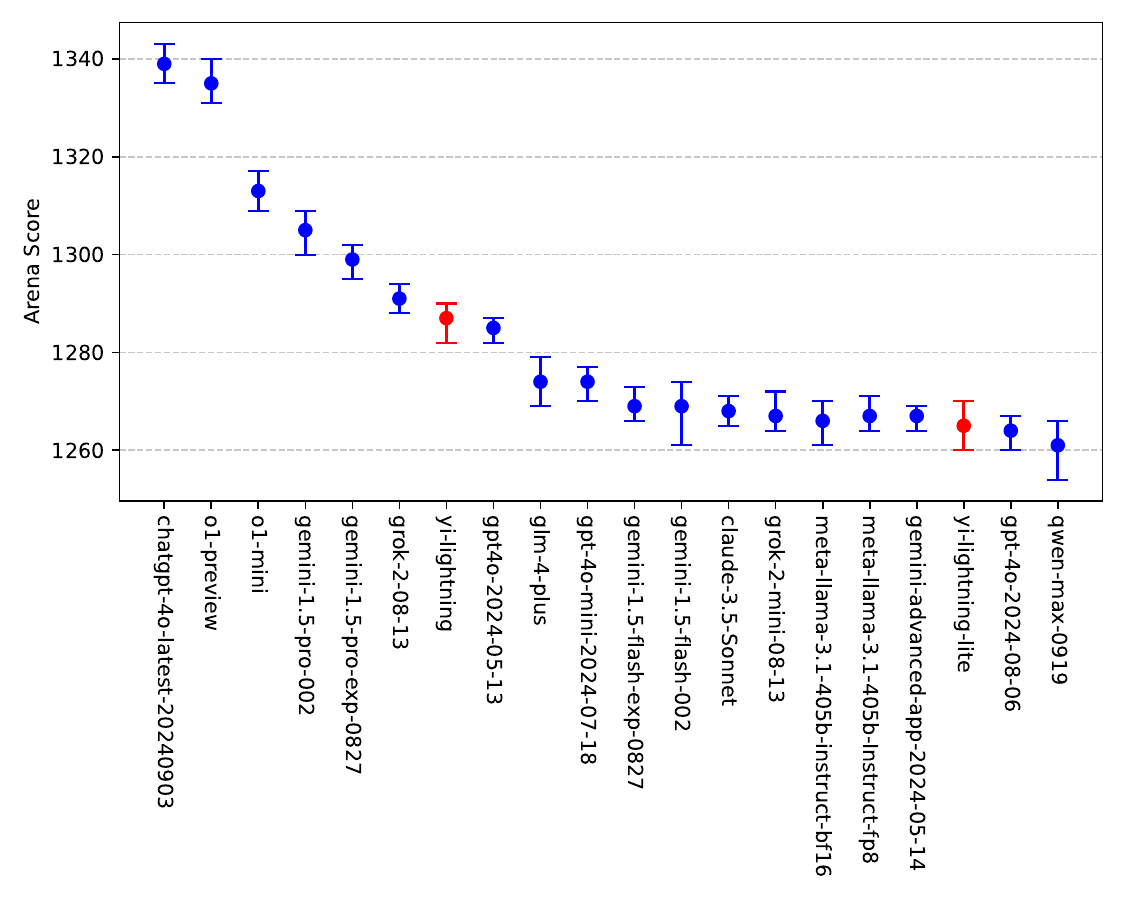}
    \caption{Snapshot of the Chatbot Arena leaderboard on October 16, 2024, with Yi-Lightning's initial appearance.
    According to the official Chatbot Arena leaderboard, Yi-Lightning ranked \textbf{6th} overall, tied with Grok-2-08-13.}
    \label{fig:chatbot_arena}
\end{figure}

Large language models (LLMs) have revealed fascinating prospects toward artificial general intelligence (AGI), attracting an enduring enthusiasm and interest of the community \citep{gpt4, team2023gemini, llama3, yang2024qwen2, 01.AI:2024aa}.
With our mission in mind to empower the community with advanced AI technology and exceptional AI service experiences, we release this technical report and introduce our new-generation flagship model, \textbf{Yi-Lightning}. 
As of its first appearance on October 16, 2024, Yi-Lightning achieved a remarkable overall ranking of \textbf{6th} place on the Chatbot Arena leaderboard \citep{mt-bench} (as shown in Figure~\ref{fig:chatbot_arena}), a leading LLM benchmark based on real-world human judgment and comparison.
In specialized categories such as Chinese, Math, Coding, and Hard Prompts, it also ranks among the top performers (ranging from 2nd to 4th place), demonstrating a comprehensive high-performance standard in practical scenarios.

We attribute Yi-Lightning's excellent performance to our innovations in model architecture (\S~\ref{sec:model_arch}), training strategies, data engineering (\S~\ref{sec:pretraining} and \S~\ref{sec:posttraining}), and infrastructure (\S~\ref{sec:infra}).
These efforts work closely together, consequently enabling Yi-Lightning's efficient training, deployment, inference, and its high practical efficacy:
\begin{itemize}[itemsep=2mm,leftmargin=6mm]
\item In terms of \textbf{model architecture}, Yi-Lightning is based on an improved mixture of experts architecture.
It employs fine-grained expert segmentation, complemented by a balanced expert routing strategy and cross-layer KV cache sharing design, achieving more efficient training and inference.
\item Regarding \textbf{training strategies}, Yi-Lightning extensively utilizes specialized multi-stage and strategic training recipes in pre-training, supervised fine-tuning, reward modeling, and human preference alignment optimization, achieving more efficient performance optimization.
\item In \textbf{data engineering}, building upon general domain foundations, we design data synthesis approaches for difficult and complex tasks (such as mathematics and coding), significantly enhancing Yi-Lightning's problem-solving capabilities.
\item For \textbf{infrastructure}, we implement substantial optimizations in parallelism strategies, node management and scheduling, storage, and network communication, greatly improving goodput performance.
\item Additionally, we implemented RAISE, a four-component framework to address the \textbf{safety} concerns across Yi-Lightning's entire lifecycle from development to deployment.
\end{itemize}

Finally, we report Yi-Lightning's evaluation results on public academic benchmarks.
While Yi-Lightning still performs competitively against other top-tier LLMs, we observe a notable disparity between the evaluation results on these academic benchmarks and real-world human judgments on Chatbot Arena, which probably results from our focus on optimizing toward practical experiences rather than overly paying attention to the benchmark scores.
This observation stimulates us to reconsider the role of the current academic benchmarks in guiding more intelligent and powerful AI systems and to design alternative approaches to evaluating model performance in practical scenarios.

\section{Model Architecture}
\label{sec:model_arch}

Similar to recent large language models \citep{dai2024deepseekmoe, yang2024qwen2, mixtral, phi3}, Yi-Lightning is fundamentally built upon the Mixture-of-Experts (MoE) architecture.
Beyond this, we introduce several architectural innovations in Yi-Lightning, including fine-grained expert segmentation methodology, advanced routing strategies, and optimized key-value cache reduction techniques.

\subsection{Fine-grained Expert Segmentation}

Recent research has revealed that as dense models grow in size, their activation patterns become increasingly sparse \citep{zhang2024exploring, zhang2024relu}.
This sparsity indicates that parameters are not uniformly utilized during inference, leading to computational inefficiencies.
The Mixture-of-Experts (MoE) architecture addresses this challenge by selectively routing tokens to activate only specific neural subsets.
However, even with MoE models, the issue of sparse activations persists within individual experts, suggesting a fundamental challenge in parameter utilization efficiency.

Drawing inspiration from \citep{dai2024deepseekmoe}, we adopt a fine-grained expert segmentation strategy.
This approach involves partitioning each expert's Feed-Forward Network (FFN) into smaller functional units, simultaneously reducing intermediate hidden dimensions while increasing the number of experts activated per token.
This fine-grained segmentation facilitates more nuanced knowledge decomposition, enhances expert activation combinations, and improves overall parameter utilization efficiency.
In practice, we observed that excessive expert segmentation substantially impacted training throughput.
Consequently, we opted for a balanced approach, implementing segmentation only to the extent necessary to maintain optimal training efficiency rather than pursuing maximum segmentation for performance gains.

\subsection{Expert Routing Strategy}

Expert routing strategy plays a crucial role in optimizing training efficiency and model quality.
Following Switch-Transformer (ST) \citep{fedus2022switch}, we initially implemented a load balancing mechanism with an auxiliary loss function for $N$ experts and a batch $\mathcal{B}$ containing $|\mathcal{B}|$ tokens:
\begin{align*}
    \mathcal{L}_{\text{ST}} = \alpha_{\text{ST}}\cdot N \cdot \sum_{i=1}^{N} f_i \cdot P_i.
\end{align*}
Here, $f_i$ represents the fraction of tokens $x$ routed to each expert, and $P_i$ denotes the average routing probability allocated to expert $i$:
\begin{align*}
    f_{i}=\frac{1}{|\mathcal{B}|} \sum_{x \in \mathcal{B}} \mathbbm{1} \left[ \left(\mathop{\arg\max}_{1\le j \le N} p_j(x) \right) =i \right], \quad P_i = \frac{1}{|\mathcal{B}|} \sum_{x \in \mathcal{B}} p_i(x),
\end{align*}
where $p_i(x)$ represents the probability of token $x$ being assigned to expert $i$.
Given that $\sum_{i=1}^N f_i = 1, \sum_{i=1}^N P_i = 1$, it can be easily shown via the Lagrange multiplier method that the loss function reaches its minimum when tokens are evenly distributed, with both $f_i$ and $P_i$ approaching $1/N$ for all experts.
Therefore, this mechanism can effectively encourage uniform routing and balanced utilization across experts during training.

However, we observed that while this load balancing effectively prevents expert collapse and enhances training efficiency, the per-expert constraints prove overly restrictive even with carefully tuned $\alpha_{\text{ST}}$.
We thus propose to relax the constraints from individual experts to Expert Parallel (EP) groups (see \S~\ref{subsec:parallel}) by introducing the \textbf{$\text{EP}$ load balancing} mechanism, which optimizes the load balance within each EP group:
\begin{align}
    \mathcal{L}_{\text{EP}} =  \alpha_{\text{EP}} \cdot N^g \cdot \sum_{i=1}^{N^g} f_i^g \cdot P_i^g
\end{align}
where $N^g$ denotes the number of experts in the EP group, with group-specific calculations:
\begin{align*}
    f^g_{i}=\frac{1}{|\mathcal{B}^g|} \sum_{x \in \mathcal{B}^g} \mathbbm{1} \left[ \left(\mathop{\arg\max}_{1\le j \le N} p_j(x) \right) =i \right], \quad 
    P^g_i = \frac{1}{|\mathcal{B}^g|} \sum_{x \in \mathcal{B}^g} p_i(x),
\end{align*}
where $\mathcal{B}^g$ denotes the batch tokens allocated to the specific EP group.

Nevertheless, a significant limitation of the above load balancing approaches still remains, as they are unable to address the \textit{token dispatching imbalance} during All-to-All communication in expert parallelism.
This imbalance results in fluctuating computation and communication intensities across EP groups, reducing training efficiency.
The issue is further compounded by expert segmentation as it increases tokens that need dispatching.
To address this challenge and enable finer-grained load balancing control, we introduce \textbf{partitioned $\text{EP}$ load balancing (PEP)}, which splits experts within each EP group into smaller partitions.
This approach ensures balanced token distribution across partitions, gradually optimizing communication and computation loads. For a partition (within a group of $N^g$ experts) containing $N^p$ local experts, the loss is computed as:
\begin{equation}
    \mathcal{L}_{\text{PEP}} = \alpha_{\text{PEP}} \cdot N^p \cdot \sum_{i=1}^{N^p} f_{i}^p \cdot P_{i}^p,
\end{equation}
where $f_{i}^p$ and $P_{i}^p$ represent the token fraction and the average routing probability for expert $i$ in this group partition, respectively:
\begin{align*}
    f_{i}^p = \frac{1}{|\mathcal{B}^p|} \sum_{x \in \mathcal{B}^p} \mathbbm{1} \left[ \left(\mathop{\arg\max}_{1\le j \le N} p_j(x) \right) =i \right], \quad
    P_{i}^p = \frac{1}{|\mathcal{B}^p|} \sum_{x \in \mathcal{B}^p} p_i(x).
\end{align*}
To maintain effective load balancing, we optimize $\mathcal{L}_{\text{PEP}}$ in conjunction with $\mathcal{L}_{\text{ST}}$ and $\mathcal{L}_{\text{EP}}$.
In practice, we carefully tuned and set $\alpha_{\text{PEP}}$, $\alpha_{\text{EP}}$, and $\alpha_{\text{ST}}$ to $10^{-3}$, $10^{-4}$, and $10^{-6}$, respectively.

\subsection{KV Cache Reduction}
\label{subsec:kv-cache}

To enhance long-context processing while substantially reducing inference costs, we introduce two key architectural innovations.
First, we observed that while most attention heads primarily focus on local context, only a small subset specializes in global information processing.
Motivated by this, we implement \textbf{hybrid attention blocks} that combine three sliding window attention \citep{mistral} layers with one full attention layer, effectively capturing both local patterns and global dependencies.
Second, we optimize memory utilization by \textbf{cross-layer KV cache reuse}, which shares key-value (KV) cache states between consecutive full attention layers and reduces memory requirements by half for full attention components.
These innovations collectively achieve up to 82.8\% memory reduction while maintaining model performance on long sequences.

\section{Pre-training}
\label{sec:pretraining}

We then describe the pre-training methodology of Yi-Lightning.
While building upon the experience of training our previous Yi models \citep{01.AI:2024aa}, we specifically optimize data processing and composition, closely integrating these improvements with pre-training progress to design an advanced multi-stage training approach.

\subsection{Data}

Our pre-training corpus comprises multilingual web documents (crawled through early 2024), books, academic papers, codebases, and question-answer pairs.
Building upon the data processing pipeline from \citep{01.AI:2024aa}, we strengthen our filtering mechanisms for unsafe content and personally identifiable information (PII), and particularly implement several key improvements detailed below.

\paragraph{Tokenization}
We employ byte-pair encoding (BPE) for text tokenization \citep{shibata1999byte} with the SentencePiece implementation \citep{kudo2018sentencepiece}.
In contrast to our previous Yi models \citep{01.AI:2024aa}, we expand the vocabulary size to 100,352 tokens to enhance multilingual support.
To improve the model's comprehension of numerical information, we decompose numbers into individual digits.
Additionally, our tokenization strategy incorporates unicode-byte encoding as a fallback mechanism for rare characters, ensuring robust fault tolerance in text processing.

\paragraph{Enhanced Mathematical and Programming Content}
We have increased the proportion of mathematical and programming content in our pre-training corpus.
Mathematical content is collected from Common Crawl using an iterative classification approach \citep{shao2024deepseekmath}, supplemented with mathematical materials from books and academic papers.
For programming content, we primarily utilize GitHub repositories, following cleaning procedures similar to \citep{guo2024deepseekcoder}.
To prevent data contamination in subsequent evaluations, we filter out entries sharing any 30-gram with the training or test sets of popular benchmarks, such as MATH \citep{math}, GSM8K \citep{gsm8k}, HumanEval \citep{humaneval}, and MBPP \citep{mbpp}.

\paragraph{Semantic-based Document Organization}
Inspired by \citet{shi2024incontextpretraining}, we implement large-scale clustering of documents with similar semantic features and concatenate them into extended sequences.
These sequences are segmented into fixed-length pieces (8,192 tokens) for pre-training, with a high-quality subset reserved for subsequent long-context extension training.

\paragraph{Fine-grained Content Classification}
We develop a series of fine-grained classifiers for text types and topics, trained on annotations generated by smaller Yi models.
The final pre-training data composition was determined through extensive experimentation with various dataset weighting schemes.
We observed that focused training on a smaller volume of high-quality domain-specific data can enhance key model capabilities.

\subsection{Training Strategy}

Our training methodology follows a three-stage approach that optimizes learning rate schedules and strategic data sampling to maximize model performance \citep{ibrahim2024simple, hu2024minicpm}: initial pre-training, mid-training, and fast-decay training.

The \textbf{initial pre-training} stage employs a warm-up schedule where the learning rate decays to half of its peak value.
This strategy enables thorough exploration of the parameter space while avoiding premature convergence.
During this stage, we emphasize data diversity to establish robust foundational capabilities across diverse domains.

In the \textbf{mid-training} stage, we focus on enhancing model capabilities and extending context length through gradual data distribution shifts.
We implement an incremental upsampling strategy for high-quality data, emphasizing complex reasoning and multilingual capabilities for low-resource languages.
We optimize training efficiency and improve throughput through dynamic batch size adjustments based on loss values.

The final \textbf{fast-decay training} stage, consuming about 12.5\% of total training tokens, combines an aggressive learning rate decay with dynamic batch size optimization.
This stage intensifies high-quality data upsampling and incorporates early instruction-tuning adaptation.
It is notably designed to be iteratively flexible, allowing multiple optimization cycles tailored to specific deployment requirements for better practical utility.

\subsection{Long Context Extension}
\label{subsec:long-context}

After the fast-decay training stage, we apply additional long-context training to extend the context window to 64K tokens.
We employ Rotary Position Embedding (RoPE) \citep{su2021roformer} with increased base frequency during extension \citep{xiong2023effective}.
This training process systematically upsamples sequences from multiple length intervals (8K-16K, 16K-32K, and 32K-64K tokens) while maintaining consistent data distribution \citep{fu2024data}.
We found that a training corpus of 20B tokens successfully developed robust long-context capabilities without compromising performance on standard benchmarks.

\section{Post-training}
\label{sec:posttraining}

We next detail Yi-Lightning's post-training methodology.
Our approach incorporates the sequential stages of Supervised Fine-Tuning (SFT) and Reinforcement Learning from Human Feedback (RLHF).
We particularly elaborate on our strategies for multi-stage model training as well as data curation and synthesis.

\subsection{Supervised Fine-tuning}
\label{subsec:sft}

\subsubsection{Data and Training Strategy}

\paragraph{Multi-stage Training and Data Curation}
Our supervised fine-tuning (SFT) process involves two sequential stages, leveraging 1.3M and 300,000 samples respectively.
The first stage focuses on enhancing fundamental capabilities in mathematics and coding through extensive synthetic data, while the second stage utilizes diverse, high-quality general-domain data to boost instruction following and problem-solving capabilities.
Furthermore, we implement a \textit{small-to-large data scaling} strategy to expand our dataset systematically across both stages.
We first compile a comprehensive prompt set from diverse sources using efficient selection strategies, and then generate corresponding responses through manual curation and synthetic approaches.
For example, in the second stage, we methodically expanded from approximately 10,000 initial, high-quality seed samples to the target of 300,000 samples.
This two-stage methodology, combined with the progressive data expansion strategy, effectively addresses data imbalance while rapidly enhancing model capabilities.

\paragraph{Synthetic Data Generation}
Synthetic data has proved instrumental, particularly for complex tasks including instruction following, code generation, and mathematical problem-solving.
We employ multiple synthesis techniques including document augmentation, self-evolution, and language translation for prompt generation.
For general tasks, we leverage multiple advanced models for response generation, combining automated systems and manual verification for quality control.
For complex tasks like coding and mathematics, we integrate search algorithms, including Monte Carlo Tree Search (MCTS) and Depth-First Search (DFS), with specialized outcome and process reward models \citep{lightman2023let} to generate diverse, accurate solutions.
These methodologies yield a substantial corpus of high-quality, diverse training data, contributing significantly to our model's initial capabilities.

\subsubsection{Optimized Implementations}

We implement \textbf{sample packing}, which concatenates multiple samples into single sequences rather than applying individual padding.
While this approach substantially reduces training sequences and improves efficiency, it can create artificial multi-turn contexts that potentially compromise certain model capabilities, particularly in multi-turn dialogues.
We address this challenge by implementing \textbf{block causal attention (BCA)}, which isolates samples within sequences through masking matrices.
We also observed that sample packing could introduce potential biases where longer samples disproportionately influence the total loss, potentially undermining short-sample task performance.
We thus develop a \textbf{sample reweighting} mechanism that equalizes loss weights across all samples within a batch, which effectively mitigates the length-induced optimization bias.

\subsection{Reinforcement Learning from Human Feedback}
\label{subsec:rlhf}

Training language models with human feedback has emerged as a crucial step for practical applications to align model behavior with human preferences \citep{bai2022training, ouyang2022training, click, rafailov2023direct, iterative-rlhf, llm-extrapolation}, which is also key to Yi-Lightning's exceptional performance on Chatbot Arena, where evaluations are based on direct human comparisons and judgments.
We next present a detailed discussion of our methodology for reward modeling (\S~\ref{subsubsec:rm}), preference data construction (\S~\ref{subsubsec:preference_data}), and alignment training (\S~\ref{subsubsec:dpo}) procedures.

\subsubsection{Reward Modeling}
\label{subsubsec:rm}

Following \cite{bai2022training, Bai:2023aa}, we implement a two-stage approach for reward model training: \textbf{preference model pre-training (PMP)} and \textbf{human-feedback fine-tuning (HFFT)}.

\paragraph{PMP Data Construction}
Our PMP dataset incorporates diverse preference datasets from public sources \citep{cui2023ultrafeedback, wang2024helpsteer2, DPO-En-Zh-20k}.
Given the varying quality standards and potential redundancies in these datasets, we implement rigorous cleaning and preprocessing protocols.
We assess dataset quality by training individual reward models and evaluating their performance on our in-house benchmarks, retaining only those datasets yielding high benchmark performance.

\paragraph{HFFT Data Construction}
The HFFT dataset is constructed using comprehensive human annotations.
We collect prompts from curated public datasets and then generate responses using model checkpoints from various SFT training phases.
These responses are evaluated across category-specific dimensions.
For instance, responses for coding prompts are assessed on instruction adherence, correctness, and code style.
We form preference pairs by selecting the highest and lowest-scoring responses for each prompt, based on weighted dimensional scores, where the pairs with insufficient score differentials are excluded.

\paragraph{Reward Model Training}
With the above curated datasets, the reward model training initializes from a pre-trained model and proceeds through the two sequential stages (PMP and HFFT) using the Bradley-Terry loss \citep{Bradley1952RankAO}.

\subsubsection{Preference Data}
\label{subsubsec:preference_data}

\paragraph{Prompts}
We generate prompts for preference learning through two approaches: collecting from public datasets and prompt synthesis.
We first collect diverse prompts from various domains (coding, mathematics, general QA, etc.) to establish a comprehensive foundation across multiple instruction contexts.
To enhance the model's capability in handling complex queries, we synthesize additional challenging prompts.
We assign complexity scores based on instruction context complexity, analytical requirements, and output format specifications.
High-scoring prompts are selected as \textit{seed prompts} and paired with diverse \textit{seed contexts} collected from high-quality web sources to create synthesized prompts.
Finally, we apply multiple deduplication techniques, including n-gram similarity analysis, embedding-based comparisons, and random downsampling, to ensure the uniqueness of prompts while preserving their diversity.

\paragraph{Preference Pairs}
To ensure balanced data distribution and rational reward assignment, we categorize prompts across multiple dimensions: complexity level, user intent clarity, and domains (e.g., math, code, general QA).
This categorization guides prompt balance adjustments and informs rating criteria weights for different categories.
For each prompt, we generate multiple responses using the SFT model with varying temperature settings.
These responses are evaluated using the reward model in \S~\ref{subsubsec:rm}, and preference pairs are formed by selecting the highest and lowest-scoring responses while ensuring a sufficient reward gap to minimize the impact of reward modeling error.

\subsubsection{Direct Preference Optimization}
\label{subsubsec:dpo}

We conduct training for human preference alignment via the direct preference optimization (DPO) algorithm~\citep{rafailov2023direct}.
Inspired by recent work on iterative DPO training~\citep{yang2024qwen2, iterative-rlhf}, we conduct the DPO training in two sequential stages: offline and online training.
In the \textbf{offline DPO training} stage, we train the model on the preference dataset constructed in \S~\ref{subsubsec:preference_data}. %
In the \textbf{online DPO training} stage, we further extend the offline training with the real-time dataset generated by the most recent model. %
For each prompt, we sample 16 candidate responses and form a preference pair using the reward model in \S~\ref{subsubsec:rm}, which is then used for model training in the next iteration.
We conducted two iterations of online DPO training in total.

We enhance the training efficiency of the conventional DPO implementation through two key optimizations.
First, instead of keeping the reference model loaded in GPU memory during training, we pre-compute and cache the log probability of samples from the preference dataset before each training iteration.
These numerical values can then be fast indexed during training without the extra overhead of loading the reference model.
Second, we leverage the fact that preference pairs share the same context to optimize computation.
For each batch of preference pairs, we first process all positive samples followed by negative samples, reusing the KV-cache of their shared contexts.
This optimization is particularly effective for long-context samples as it eliminates redundant context processing.
Consequently, the two improvements substantially reduce GPU memory usage and enhance overall training efficiency.

\section{Infrastructure}
\label{sec:infra}

\subsection{Parallelism Optimization}
\label{subsec:parallel}

Given the architectural characteristics of MoE models (\S~\ref{sec:model_arch}), we implement a hybrid parallelization strategy combining \textbf{expert parallelism} and \textbf{pipeline parallelism}.
We further enhance the pipeline parallelism through several optimizations, including customized pipeline stage partitioning and fine-grained gradient recomputation strategies.
These improvements enable optimal memory utilization and workload distribution across devices while maintaining training stability and enhancing overall throughput.

To fully exploit the advantages of both hybrid attention (\S~\ref{subsec:kv-cache}) and context parallelism in long-context scenarios (\S~\ref{subsec:long-context}), we introduce several refinements to the context parallelism implementation.
These modifications enable efficient integration with the hybrid attention mechanism, particularly in optimizing the distribution of sliding window attention computations across the context parallel dimension.
Our approach significantly reduces the computational burden on individual context parallel ranks, resulting in an up to 70\% training speedup.

\subsection{Inference Optimization}

Yi-Lightning leverages a high-performance inference engine optimized specifically for LLM inference, effectively addressing the computational and memory bottlenecks. 
Through integrated algorithmic and engineering optimizations, the system achieves substantial reductions in resource consumption while delivering exceptional inference efficiency.
The key optimizations include:

\paragraph{Advanced Asynchronous Scheduling at Engine Level}
Traditional LLM inference solutions often suffer from suboptimal GPU utilization (typically below 70\%) due to serial dependencies between modules causing GPU idle time.
We implement sophisticated multi-module, multi-process asynchronous scheduling that decouples task execution and minimizes inter-module latency.
This enhancement achieves 95\% GPU utilization in high-concurrency scenarios, markedly improving both engine performance and hardware resource efficiency.

\paragraph{Optimized FP8 Quantization and Hardware-Aware Operator Design}
Yi-Lightning's architecture is fundamentally designed with GPU hardware characteristics in mind, particularly for FP8 quantization compatibility.
The model architecture precisely aligns with hardware specifications, maintaining algorithmic precision while maximizing hardware utilization. 
Our training infrastructure fully exploits the Nvidia Hopper architecture through custom-developed high-performance operators.
A notable example is our implementation of the Mixture-of-Experts (MoE) operator, which employs an expert-parallel strategy achieving 1,200 TFLOPS per card at FP8 precision on Hopper GPUs. This represents a performance improvement exceeding 100\% for operator execution, substantially enhancing overall inference efficiency.

The combined impact of these optimizations - enhanced hardware utilization through asynchronous scheduling and efficient operator implementation - enables Yi-Lightning to effectively address computational and memory constraints in high-concurrency, high-throughput inference scenarios, making it ideally suited for large-scale AI service deployment.

\subsection{Goodput Optimization}

Industry leaders like Meta \citep{llama3}, Google \citep{team2023gemini}, and Alibaba \citep{dong2024boosting} have achieved goodput levels exceeding 90\% by employing advanced techniques such as fast checkpointing, fault-tolerant scheduling, and hardware redundancy.
For instance, Meta's Grand Teton AI system~\citep{Bjorlin_Bjorlin_2022} leverages RDMA over RoCE and Infiniband networks to maintain high performance, while ByteDance's MegaScale \citep{jiang2024megascale} system focuses on rapid error detection and recovery to minimize downtime. Similarly, Google and Alibaba have implemented proactive hardware health monitoring and network optimizations to sustain near-maximum goodput even in the face of frequent hardware failures.

Building on these insights, we adopt a multi-layered approach to goodput optimization  in our large-scale GPU cluster, \textbf{XCloud}:

\paragraph{Fault Tolerance through Proactive and Reactive Mechanisms}
One of the most significant contributions to goodput optimization is the combination of proactive and reactive fault discovery strategies. Proactive measures such as routine, entrance, and preflight tests ensure cluster health by identifying potential hardware and software issues before they impact workloads. On the reactive side, XCloud employs advanced monitoring tools like node exporters\footnote{\url{https://github.com/prometheus/node_exporter}} and custom InfiniBand metrics collectors to detect faults in real time. These systems work in tandem to minimize the duration and impact of failures, enabling rapid recovery and reducing wasted computational resources. This dual-layer approach ensures that GPU resources remain optimally utilized even in the face of frequent hardware or network failures.

\paragraph{Memory-Based Asynchronous Checkpointing}
Traditional checkpointing systems, like those relying on distributed file systems (e.g., GPFS~\citep{gpfs}), often introduce significant overhead, leading to idle GPU time during save operations. XCloud's memory-based asynchronous checkpointing drastically reduces the time required to save model states, from several minutes to just 3--5 seconds. This innovation not only minimizes the GPU idle period but also encourages more frequent checkpointing, reducing computational waste during recovery. The result is a substantial boost in system resilience and overall efficiency, contributing directly to achieving and maintaining goodput levels above 99\%.

\section{Safety}
\label{sec:safety}

\begin{figure}[htbp]
    \centering
    \vspace{1mm}
    \includegraphics[width=0.95\linewidth]{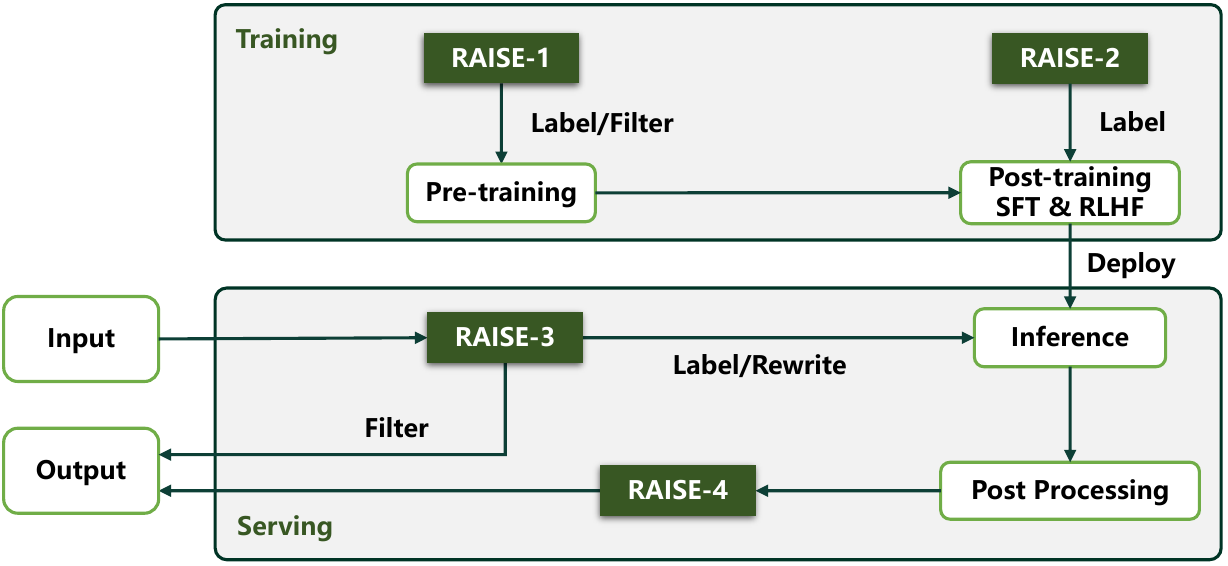}
    \vspace{2mm}
    \caption{Illustration of RAISE (Responsible AI Safety Engine) system.}
    \label{fig:safe-handling}
\end{figure}

As large language models continue to grow in capability, it is crucial to ensure their safe and responsible operation across varying and complex scenarios \citep{gpt4, llama3, zou2023universal, wei2023jailbroken, llm-safeguard, wang-etal-2024-boosting-llm, zhao2024diver}.
To address the safety concerns of Yi-Lightning, we develop \textbf{RAISE (Responsible AI Safety Engine)}, a comprehensive safety framework illustrated in Figure~\ref{fig:safe-handling}.
RAISE is designed to provide robust safety capabilities throughout the model's entire lifecycle, from development to deployment, effectively minimizing potential risks and threats through systematic safety mechanisms.

The RAISE framework comprises four integral components (RAISE-1 to RAISE-4), corresponding to pre-training, post-training, and inference-time input/output processing.
Through sophisticated technical approaches and their synergistic integration, these components collectively ensure model safety while maintaining optimal user experience. %

\paragraph{RAISE-1: Pre-training Safety}
At the pre-training phase, we implement a safety model for pre-training data filtration. 
We develop classification models based on Transformer and DNN architectures, trained on high-quality compliant datasets.
These models form an evaluation and filtering pipeline for the pre-training corpus, ensuring data reliability, minimizing erroneous information and biased content, preventing privacy data leakage, and enhancing model safety and compliance.

\paragraph{RAISE-2: Post-training Optimization}
During post-training, we implement fine-tuning strategies to optimize safety performance across different application scenarios.
Our approach incorporates evaluation and scoring mechanisms during SFT and RLHF stages, using reward engineering to encourage safe responses and penalize potentially harmful outputs.
The additional quality control processes further ensure proper value alignment while maintaining core model performance.

\paragraph{RAISE-3: Input Safety Processing}
For inference-time input processing, we deploy safety assessment mechanisms that analyze and filter content.
The system identifies potentially harmful content, including malicious, discriminatory, or hateful elements, while ensuring input safety and compliance.
These mechanisms minimize risks of the model being manipulated while maintaining performance under various input conditions.

\paragraph{RAISE-4: Output Safety Control}
The output safety control system implements real-time detection and optimization across key dimensions: value alignment, bias detection, legal compliance, accuracy assessment, and content appropriateness.
This component integrates safety mechanisms to ensure output quality while maintaining efficiency, balancing safety requirements with response speed.

Through this framework, RAISE provides a foundation for responsible AI development and deployment, ensuring safety across Yi-Lightning's lifecycle while maintaining performance and user satisfaction.
The interaction between these components creates a safety ecosystem addressing both current and emerging challenges.

\section{Evaluation}
\label{sec:evaluations}

\paragraph{Chatbot Arena}

As shown in Figure~\ref{fig:chatbot_arena}, with the initial appearance on Chatbot Arena\footnote{\url{https://lmarena.ai/}} \citep{mt-bench} on October 16, 2024, our flagship model Yi-Lightning achieved a remarkable overall ranking of \textbf{6th} place (Arena score \textbf{1287}), performing on par with GPT-4o-0513 (ranked 7th, Arena score 1285).
In specialized rankings, Yi-Lightning also exhibited remarkable performance: \textbf{2nd} in \textit{Chinese}, \textbf{3rd} in \textit{Multi-Turn} and \textit{Math}, and \textbf{4th} in \textit{Coding}, \textit{Hard Prompts}, and \textit{Longer Query} categories.
Since the Chatbot Arena rankings derive from authentic human comparisons and voting, these results powerfully demonstrate Yi-Lightning's exceptional ability to fulfill user needs and its satisfactory alignment with human preferences in real-world applications.

\paragraph{Academic Benchmarks}
We report the evaluation results on several representative, public academic benchmarks: GPQA \citep{gpqa} for general knowledge, MATH \citep{math} for mathematical reasoning, HumanEval \citep{humaneval} for coding, and IFEval \citep{ifeval} for instruction following.
We also conduct the LLM-as-a-judge evaluations\footnote{We employ \textit{GPT-4o-0513} for all the LLM-as-a-judge evaluations.} on WildBench \citep{lin2024wildbench}, Arena-Hard\footnote{
Since Yi-Lightning's API serving trades off inference speed and accuracy, evaluation results based on the API serving may be slightly lower than our reported evaluation results based on local deployment.} \citep{arena-hard}, AlignBench-v1.1 \citep{alignbench}, and MT-Bench \citep{mt-bench}.

We present in Table~\ref{tab:comparison_open} the comparison between Yi-Lightning and the top-tier open-weight LLMs, including Qwen2.5-72B-Instruct \citep{qwen2.5}, DeepSeek-V2.5 \citep{deepseek-v2}, Mistral-Large-Instruct-2407 \citep{mistral}, and Llama3.1-70B/405B-Instruct \citep{llama3}.
We also show in Table~\ref{tab:comparison_closed} the comparison with the top-tier proprietary LLMs, including GPT-4o-0513, Claude-3.5-Sonnet-20240620, and our last-generation model Yi-Large-Preview.
Overall, Yi-Lightning remains competitive on these academic benchmarks.

\begin{table}[ht]
\centering
\caption{Comparison with the top-tier open-weight LLMs on public academic benchmarks.}
\vspace{2mm}
\adjustbox{center=\textwidth}{
\begin{tabular}{lcccccc}
\toprule
& \tabincell{c}{Qwen2.5 \vspace{0.5mm} \\ 72B \vspace{0.5mm} \\ Instruct} & \tabincell{c}{DeepSeek \vspace{0.5mm} \\ V2.5} & \tabincell{c}{Mistral \vspace{0.5mm} \\ Large \vspace{0.5mm} \\ Instruct-2407} & \tabincell{c}{Llama3.1 \vspace{0.5mm} \\ 70B \vspace{0.5mm} \\ Instruct} & \tabincell{c}{Llama3.1 \vspace{0.5mm} \\ 405B \vspace{0.5mm} \\ Instruct-FP8} & \tabincell{c}{\textbf{Yi-Lightning}} \\
\midrule
\tabincell{l}{GPQA \vspace{-1mm}\\ \scriptsize{\textit{0-shot}} } & 49.1 & 42.4 & 50.1 & 45.1 & \textbf{53.0} & 50.9 \vspace{1mm} \\
\tabincell{l}{MATH \vspace{-1mm}\\ \scriptsize{\textit{0-shot}} } & \textbf{82.7} & 73.9 & 73.3 & 67.1 & 67.7 & 76.4 \vspace{1mm} \\
\tabincell{l}{HumanEval \vspace{-1mm}\\ \scriptsize{\textit{0-shot}} } & \textbf{86.0} & 85.4 & \textbf{86.0} & 76.2 & 84.1 & 83.5 \vspace{1mm} \\
\tabincell{l}{IFEval \vspace{-1mm}\\ \scriptsize{\textit{Prompt Loose}} } & 86.0 & 82.1 & 82.8 & 87.1 & \textbf{88.5} & 81.9 \vspace{1mm} \\
\tabincell{l}{WildBench \vspace{-1mm}\\ \scriptsize{\textit{Judge: GPT-4o-0513}} }  & 59.9  & 57.9 & 57.4 & 49.0 & 51.6 & \textbf{65.1} \vspace{1mm} \\ 
\tabincell{l}{Arena-Hard \vspace{-1mm}\\ \scriptsize{\textit{Judge: GPT-4o-0513}} } & 90.5 & 88.3 & 85.1 & 74.0 & 71.2 & \textbf{91.8} \vspace{1mm} \\
\tabincell{l}{AlignBench-v1.1 \vspace{-1mm}\\ \scriptsize{\textit{Judge: GPT-4o-0513}} } & 7.51 & 7.38 & 7.10 & 5.81 & 5.56 & \textbf{7.54} \vspace{1mm} \\
\tabincell{l}{MT-Bench \vspace{-1mm}\\ \scriptsize{\textit{Judge: GPT-4o-0513}} } & 8.62 & 8.43 & 8.53 & 8.23 & 8.36 & \textbf{8.75} \\
\bottomrule
\end{tabular}
}
\label{tab:comparison_open}
\end{table}

\begin{table}[ht]
    \centering
    \caption{Comparison with the top-tier proprietary LLMs (GPT-4o-0513 and Claude-3.5-Sonnet-20240620) and our previous-generation model (Yi-Large-Preview) on public academic benchmarks.}
    \vspace{2mm}
    \adjustbox{center=\textwidth}{
    \begin{tabular}{lccccc}
    \toprule
    & \tabincell{c}{Yi-Large \vspace{0.5mm} \\ Preview} & \tabincell{c}{GPT-4o \vspace{0.5mm} \\ 0513} & \tabincell{c}{Claude-3.5-Sonnet \vspace{0.5mm} \\ 20240620} & \tabincell{c}{\textbf{Yi-Lightning}} \\
    \midrule
    \tabincell{l}{GPQA \vspace{-1mm}\\ \scriptsize{\textit{0-shot}} } & 43.8 &  51.9 & \textbf{57.8} & 50.9 \vspace{1mm} \\
    \tabincell{l}{MATH \vspace{-1mm}\\ \scriptsize{\textit{0-shot}} } & 62.6 &  76.0 & 74.0 & \textbf{76.4} \vspace{1mm} \\
    \tabincell{l}{HumanEval \vspace{-1mm}\\ \scriptsize{\textit{0-shot}} } & 75.6  & \textbf{90.5} & 88.3 & 83.5 \vspace{1mm} \\
    \tabincell{l}{IFEval \vspace{-1mm}\\ \scriptsize{\textit{Prompt Loose}} } & 79.3  & 87.6 & \textbf{88.5} & 81.9 \vspace{1mm} \\
    \tabincell{l}{WildBench \vspace{-1mm}\\ \scriptsize{\textit{Judge: GPT-4o-0513}} } & 55.3 & 59.3 & 54.7 & \textbf{65.1} \vspace{1mm} \\
    \tabincell{l}{Arena-Hard \vspace{-1mm}\\ \scriptsize{\textit{Judge: GPT-4o-0513}} } & 79.1  & \textbf{92.9} & 85.6 & 91.8 \vspace{1mm} \\
    \tabincell{l}{AlignBench-v1.1 \vspace{-1mm}\\ \scriptsize{\textit{Judge: GPT-4o-0513}} } & 7.20  & \textbf{7.59} & 7.17 & 7.54 \vspace{1mm} \\
    \tabincell{l}{MT-Bench \vspace{-1mm}\\ \scriptsize{\textit{Judge: GPT-4o-0513}} } &  8.32  & 8.59 & 6.96 & \textbf{8.75} \\
    \bottomrule
    \end{tabular}
    }
    \label{tab:comparison_closed}
\end{table}

\subsection*{Final Discussion}
Finally, we discuss our observed disparity between open-weight and proprietary models' performance on public academic benchmarks and real-world user preferences (as reflected in the Chatbot Arena rankings).
This probably results from the fact that our development process paid more attention to our in-house human assessment experience, rather than overly focusing on academic benchmark scores.
For instance, when conducting math-specific model training (e.g., in \S~\ref{subsec:sft}), we did not strictly restrict the model's output format (e.g., ending with ``$\mathrm{The\ final\ answer\ is}\ \backslash\mathrm{boxed}\{...\}$''), as we believe that constraining the model's output content or format might harm its generation diversity, thereby implicitly impacting optimization effectiveness and user experience.
These evaluation results prompt us to rethink and reassess the role of public academic benchmarks in guiding the development of more intelligent and powerful AI systems.

\appendix
\vspace{10mm}
\section{Author List and Contributions}

We list our team members in alphabetical order.
All authors contribute equally to this work. 

\subsection*{Pre-training}

\begin{multicols}{3}
\begin{itemize}
\item Bei Chen
\item Chao Li
\item Chengen Huang
\item Fan Zhou
\item Ge Zhang
\item Jun Tian
\item Peng Liu
\item Shiming Yang
\item Wenhao Huang
\item Xiaoyi Ren
\item Xinyao Niu
\item Yanpeng Li
\item Yuchi Xu
\item Zhiyuan Liu
\end{itemize}
\end{multicols}

\subsection*{Post-training}

\begin{multicols}{3}
\begin{itemize}
\item C.X. Lv
\item Chenglin Cai
\item Chujie Zheng
\item Guoyin Wang
\item Heng Ji
\item Katherine Su
\item Ming Song
\item Shawn Wang
\item Tianhang Zhu
\item Xiang He
\item Xiaohui Hu
\item Yuxuan Sha
\end{itemize}
\end{multicols}

\subsection*{Infrastructure}

\begin{multicols}{3}
\begin{itemize}
\item Feng Hu
\item Jiangcheng Zhu
\item Lihuan Zhang
\item Liying Li
\item Mou Li
\item Qicheng Hu
\item Wen Xie
\item Xiaobo Chen
\item Yongke Zhao
\item Zhaodong Yan
\item Zirui Zhang
\item Zonghong Dai
\end{itemize}
\end{multicols}

\subsection*{Safety}

\begin{multicols}{3}
\begin{itemize}
\item Shijun Zhou
\item Shiyong Li
\item Yongzhen Luo
\end{itemize}
\end{multicols}

\subsection*{Evaluation}

\begin{multicols}{3}
\begin{itemize}
\item Alan Wake
\item Daniel Cooper
\item Howard Qiu
\end{itemize}
\end{multicols}

\clearpage
\bibliographystyle{plainnat}
\bibliography{references}

\end{document}